\pdfoutput=1

\documentclass[11pt]{article}

\usepackage{acl}

\usepackage{times}
\usepackage{latexsym}
\usepackage{graphicx}

\usepackage[T1]{fontenc}

\usepackage[utf8]{inputenc}

\usepackage{microtype}

\title{The Right Model for the Job: An Evaluation of Legal Multi-Label Classification Baselines}

\author{
Martina Forster \and
Claudia Schulz \and
Prudhvi Nokku \and \\
{\bf Melicaalsadat Mirsafian} \and 
{\bf Jaykumar Kasundra} \and
{\bf Stavroula Skylaki} \\
Thomson Reuters Labs
}

\begin{document}
\maketitle
\begin{abstract}
Multi-Label Classification (MLC) is a common task in the legal domain, where more than one label may be assigned to a legal document. A wide range of methods can be applied, ranging from traditional ML approaches to the latest Transformer-based architectures. In this work, we perform an evaluation of different MLC methods using two public legal datasets, POSTURE50K and EURLEX57K. By varying the amount of training data and the number of labels, we explore the comparative advantage offered by different approaches in relation to the dataset properties. Our findings highlight DistilRoBERTa and LegalBERT as performing consistently well in legal MLC with reasonable computational demands. T5 also demonstrates comparable performance while offering advantages as a generative model in the presence of changing label sets. Finally, we show that the CrossEncoder exhibits potential for notable macro-F1 score improvements, albeit with increased computational costs. 
\end{abstract}

\section{Introduction}
Multi-label classification (MLC) is a common task in the legal domain, where more than one label may be assigned to a legal document \cite{chalkidis-etal-2022-lexglue}. One of the challenges of legal MLC is the long-tail label distribution, i.e., few labels occur frequently, whereas the majority of labels occurs very rarely in the data \citep{chalkidis-etal-2019-extreme}. 

In the legal domain, a number of studies have explored the performance of various MLC methods, ranging from traditional ML approaches based on sparse vectors such as TF-IDF \citep{Xiao2018} to the latest state-of-the-art Transformer based models \citep{Kementchedjhieva2023}. Additionally, domain-specific models trained on legal text may exhibit enhanced performance compared to general-purpose models \citep{Chalkidis2020}.  

For practical MLC applications, the choice of an appropriate method depends on the data properties such as dataset size,  number of labels and stable vs. changing label set, the model performance, and technical considerations, such as resource availability, cost, and runtime. It is therefore important for applied researchers to rapidly establish an effective experimentation strategy by examining a set of appropriate baseline models. 

We aim to provide an informative starting collection of baselines by comparing several established MLC approaches. To elucidate the advantages and weaknesses of the different approaches, we anaylse: 1) the effect of label quantity on the performance of the various methods, 2) the effect of dataset size on the performance of the various methods, 3) the performance of domain-specific vs generic models, 4) the time, cost, and performance trade-offs of different methods.

\section{Related Work}

In the legal domain, MLC is a common task with a broad range of applications such as legal motion detection \citep{Vacek2019}, case outcome prediction \citep{Medvedeva2020}, EU legislative documents categorization \citep{chalkidis-etal-2019-eurlex57k}, user right violations in terms of services \citep{lippi-etal-2019}, and legal procedural posture classification \citep{Song2022}. 

Especially in real-world legal scenarios, MLC tasks are often challenging due to high label imbalance as well as patterns of label co-occurrences. In the legal domain, this phenomenon is exacerbated in the presence of extremely large label sets \citep{chalkidis-etal-2019-extreme}, incomplete or unreliable annotation data \citep{braun2023}, and scarcity of voluminous and high quality datasets \citep{Xiao2018, Chalkidis2019, Chalkidis2021, Song2022}. Legal documents present additional challenges in the form of domain specific vocabulary,
which makes domain-adapted Transformer-based language models outperform out-of-the-box variants \citep{Chalkidis2020}.

There are several machine learning methods that can be adopted for MLC problems. Early approaches were based on TF-IDF vectors \citep{Xiao2018}, while recent systems increasingly rely on Transformers, either encoder-only approaches, such as legal-domain BERT \citep{Chalkidis2020, Henderson2022, casehold-custom-legalbert} and RoBERTa \citep{Song2022}, or encoder-decoder architectures such as T5 \citep{Liu2021, Kementchedjhieva2023}. In addition, loss functions specifically designed to address the imbalanced label distributions, such as focal loss \citep{Lin2017}, may be used to improve the training process. Finally, the formulation of the MLC task as a sentence similarity task between text and label using SBERT-based approaches \citep{Reimers2019} may be applied in the case of frequently evolving label sets.

\section{Experiments}
To study the effect of data set size and label quantity on the performance of different MLC algorithms, we use two legal MLC datasets to simulate various data scenarios, POSTURE50K \citep{Song2022} and 
EURLEX57K \citep{chalkidis-etal-2019-eurlex57k} (for detailed dataset description see \ref{appendix:datasets}).

\subsection{Dataset Construction}

For model training, we retain the original train-dev-test splits of the dataset and take data samples from within each split, i.e., original train data will never be part of the test data in any of our data samples.

To investigate the effect of label quantity on model performance, we experiment with using only the $k$ most frequent labels for $k \in [5, 20, 50, 100, 200, 1000]$\footnote{$1000$ is only used for EURLEX57K since POSTURE50K has less than 1000 labels.} in the train, dev, and test sets. This is done by removing labels not in the $k$ most frequent ones from each data point and discarding data points without labels.
The effect of dataset size on model performance is accounted for by downsampling the number of training data points  $m$ to $m \in [1000, 2000, 5000, 10000]$ (this is done after label sampling) (for details see \ref{appendix:sampling}).

The different choices of $k$ and $m$ result in 20 sampled training sets for POSTURE50K and 24 for EURLEX57K.

\subsection{MLC Models}

We choose three sparse vector similarity methods, namely ClassTFIDF, DocTFIDF, and BM25 (details can be found in \ref{appendix:sparse}). We compare these with different types of Transformer architectures: We experiment with two frequently used language models fine-tuned for MLC, DistilRoBERTa \citep{Sanh2019} and the domain-specific LegalBERT \citep{casehold-custom-legalbert}, both fine-tuned using focal loss \citep{Lin2017} to mitigate the impact of label imbalances. T5 is chosen as a generative approach \citep{Raffel2020}, while BiEncoder and CrossEncoder architectures compare semantic similarity between label and text within a language model \citep{Reimers2019}. We choose LegalBERT \citep{casehold-custom-legalbert} as the model for the BiEncoder and CrossEncoder architectures.

Note that we use default parameters for all models as we aim to provide solid baselines rather than optimised solutions. Details about the model parameters can be found in \ref{appendix:hps}.
All Transformer models in our experiments use only the first 512 tokens of input text, a common way of handling longer legal documents \citep{mamakas-etal-2022-legal-longformer}.

\section{Results}
\begin{figure*}[t]
    \centering
    \includegraphics[width=\textwidth]{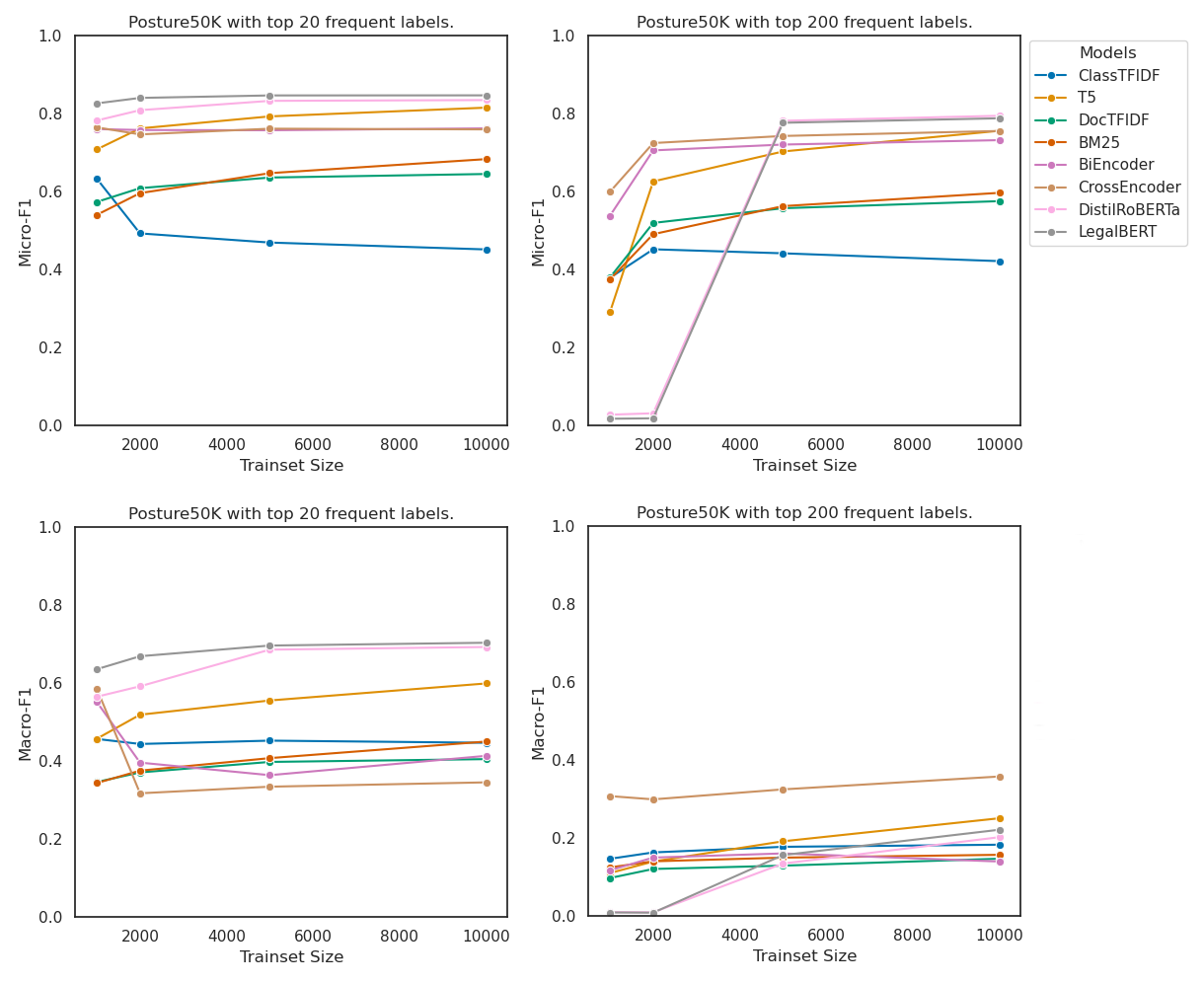}
    \caption{Micro- and macro-F1 scores of multi-label classifiers on POSTURE50K data with top 20 and top 200 labels for different training set sizes.}
    \label{fig:fig1}
\end{figure*}

To gain insights on the strengths of each model on the legal MLC datasets, we examine the performance of the different models trained on the various size trainsets. Figure~\ref{fig:fig1} shows the micro- and macro-F1 scores on the POSTURE50K dataset. Similar trends are observed on the EURLEX57K dataset, as illustrated in 
Figure~\ref{fig:eurlex57k} in~\ref{appendix:EURLEX57K}.

\subsection{What is the influence of dataset size?}
Figure~\ref{fig:fig1} shows that some models can handle a small training set better than others. 
LegalBERT and DistilRoBERTa are unable to learn at all\footnote{The F1 score is close to 0 due to precision being close to 0, whereas recall is high, indicating that the model predicts most labels for almost all data points.} when there is not enough training data compared to the number of labels, which is the case for 
2000 or less data points with 200 labels, and 5000 or less data points with 1000 labels (also see Figures~\ref{fig:eurlex57k} and~\ref{fig:eurlex1000} in~\ref{appendix:EURLEX57K}). For EURLEX57K these models are able to already learn with slightly smaller training sets. \citet{leite-etal-2020-small_data_low_performance} and \citet{iclr_small_data_low_performance} also observe that BERT models may not be able to learn with very small training sets.

All other models are less affected by a small training set, and performance
increases only slightly when training with more than 5000 data points. We observe that the macro-F1 benefits more from additional training data than the micro-F1, especially with larger numbers of labels,
indicating that models can improve performance on minority labels when 
more data becomes available.

\subsection{What is the influence of label quantity?}
We observe that the performance of all models decreases with a higher number of labels. All models show a sharp drop in performance from 5 to 20 labels and a more gradual decrease with increased label quantity thereafter (also see Figure~\ref{fig:vary_labels} in~\ref{appendix:EURLEX57K}).
TF-IDF model performance is particularly affected by the increase in labels.
Similar to dataset size, label quantity affects macro-F1 more than micro-F1.

\subsection{Are domain-specific models better?}
Comparing the performance of DistilRoBERTa and LegalBERT, we observe on-par or slightly better performance of LegalBERT for POSTURE50K, while the opposite holds for EURLEX57K. This may be because labels in POSTURE50K are from the legal domain (e.g., "On Appeal" or "Motion to Dismiss") while labels in EURLEX57K are more general (e.g., "animal product" or "tropical fruit"), thus requiring less legal understanding.

\begin{table*}[th]
\centering
\begin{tabular}{llrrrrrrrrrr}
\hline
&&&& \multicolumn{2}{c}{2k - 20} & \multicolumn{2}{c}{2k - 200} & \multicolumn{2}{c}{10k - 20} & \multicolumn{2}{c}{10k - 200}\\
\textbf{Model} & \textbf{GPUs} & \textbf{CPUs} & \textbf{GiB} & \textbf{h} & \textbf{\$} & \textbf{h} & \textbf{\$} & \textbf{h} & \textbf{\$}  & \textbf{h} & \textbf{\$}\\
\hline
 ClassTFIDF     & --                    & 4 & 16 & 0.4 & 0.1 & 0.6 & 0.2 &  0.5 & 0.1  & 0.8 & 0.3\\
 DocTFIDF       & --                    & 4 & 16 & 0.3 & 0.1 & 0.3 & 0.1 &  0.5 & 0.1  & 0.5 & 0.1\\
 BM25           & --                    & 4 & 16 & 1.5 & 0.7 & 1.6 & 0.8 &  7.4 & 4.1  & 6.8 & 3.8\\
 T5             & 1 T4          & 4 & 16 & 3.2 & 2.9 & 2.8 & 2.5 &  4.7 & 4.1  & 4.4 & 3.9\\
 BiEncoder      & 1 Tesla V100   & 8 & 61 & 0.6 & 2.3 & 0.8 & 3.3 &  2.0 & 8.1  & 2.5 & 10.0\\
 CrossEncoder   & 1 Tesla V100   & 8 & 61 & 1.2 & 7.4 & 6.1 & 44.7 & 2.6 & 12.8 & 7.6 & 50.4\\
 DistilRoBERTa  & 1 T4          & 8 & 32 & 1.5 & 1.7 & 0.6 & 0.7  & 2.4 & 2.6  & 5.6 & 5.9\\
 LegalBERT      & 1 T4          & 8 & 32 & 2.3 & 2.5 & 1.0 & 1.1 &  4.1 & 4.4  & 12.0 & 12.7\\
\hline
\end{tabular}
\caption{Time (h) and cost (\$) for training and evaluating models using different sampled training sets (2k, 10k) and numbers of labels (20, 200) of POSTURE50K and specification of compute resources used.}
\label{tab:cost_time}
\end{table*}

\subsection{What are the best legal MLC baselines?}

In Figure~\ref{fig:fig1} we can see that DistilRoBERTa and LegalBERT are the top performing algorithms for both datasets in terms of micro-F1 score, except if there is not enough training data. T5 is a close contender, and, since it is a generative model, suitable for changing label sets. It can simply be fine-tuned on a smaller dataset with new labels or even be used in a zero-shot setup, whereas RoBERTa and LegalBERT need to be trained from scratch on the full data to enable new label prediction.
The best vector similarity methods are DocTFIDF and BM25, showing reasonable performance throughout. 

In terms of macro-F1, DistilRoBERTa and LegalBERT are mostly in the top range as well. However, for larger numbers of labels, the CrossEncoder can significantly improve the macro-F1 score for POSTURE50K: when dealing with 200 labels it outperforms the other methods by a large margin. For EURLEX57K we do not observe this behaviour, with the CrossEncoder performing worse than DistilRoBERTa and LegalBERT in terms of macro-F1. This may be because the legal labels in POSTURE50K have higher semantic similarity with the legal text being classified than the more general labels in EURLEX57K. 
Similar to T5 the CrossEncoder is suitable for changing label sets, including zero-shot predictions.



\subsection{What is the cost-performance trade-off?}

Table~\ref{tab:cost_time} shows the time and cost of training and evaluating the different models when varying label quantity and training set size for POSTURE50K (similar trends are observed for EURLEX57K)
\footnote{Note that the BiEncoder and CrossEncoder were run on a more powerful, and thus faster but more expensive, GPU instance than the other Transformer models. Training time on the T4 GPU would be much slower.
Also note that training time and cost is deceptively low for DistilRoBERTa and LegalBERT with 2000 training data points and 200 labels since they are unable to learn and thus training stops early.}.

Larger training sets and label quantities increase training and evaluation time across the board. CrossEncoder, DistilRoBERTa, and LegalBERT become significantly slower with increasing number of labels, whereas the speed of T5 is not affected by more labels at all.

TF-IDF methods are significantly faster than BM25. Since DocTFIDF also has reasonable prediction performance, it has an excellent cost-performance trade-off.
The most expensive algorithm by a large margin, especially for large label sets, is the CrossEncoder, incurring 10x the cost of DistilRoBERTa for the training set with 10,000 data points and 200 labels. Whether the increased cost is worth the increased macro-F1 performance will depend on the application.
DistilRoBERTa is nearly twice as fast (and thus half as expensive) as LegalBERT due to its lower number of model parameters. Since the performance of DistilRoBERTa and LegalBERT is very similar, DistilRoBERTa has a very good cost-performance trade-off.

\section{Conclusions and Future Work}
In this work, we explore a number of established MLC approaches in the legal domain. Our aim is to provide insights for the legal community by experimenting with a collection of common baselines in relation to various data and hardware properties.


In summary, our experiments show that DistilRoBERTa and LegalBERT generally have a reasonable cost-performance trade-off for legal MLC, yielding superior results, unless the training set size is too small. T5 also gives reasonable performance, while offering flexibility in the case of changing label sets. DocTFIDF is worth trying, as it is fast and cheap to run. CrossEncoders can yield significant performance gains in macro-F1 score in some cases, although at a higher price. 

In future work, we will experiment with algorithms able to handle long documents, such as Longformer \cite{mamakas-etal-2022-legal-longformer, chalkidis-etal-2023-lexfiles, niklaus-giofre-2023-budget-longformer}
and LLMs \cite{trautmann2022legal,chatgpt-jack-of-all-trades,gema2023-llama-finetuning}.



\bibliography{bib_claudia}
\bibliographystyle{acl_natbib}

\appendix

\section{Appendix}

\subsection{Datasets}
\label{appendix:datasets} 

\subsubsection{POSTURE50K}
POSTURE50K \citep{Song2022} is a multi-label dataset consisting of 50,000 legal cases from US courts that were manually labeled with 256 legal procedural postures. On average, there are 2,901 words per document. The training set contains 31,944, the validation set 7,991, and the test set 10,065 legal cases, respectively.
To the best of our knowledge, state-of-the-art performance on the POSTURE50K dataset is 0.272 macro-F1 by LAMT\_MLC \citep{Song2022}, a custom pre-trained RoBERTa architecture with label-attention, and 0.812 micro-F1 by APLC\_XLNet \citep{aplc-xlnet}, a fine-tuned XLNet model with clustering, as reported by \citet{Song2022}.

\subsubsection{EURLEX57K}
EURLEX57K \citep{chalkidis-etal-2019-eurlex57k} comprises 57,000 English EU legislative documents from EURLEX, which were annotated with 4,271 concepts from the European Vocabulary EUROVOC. The average number of words per document is 727, and the number of documents is 45,000 for train, 6,000 for validation, and 6,000 for test, respectively. 
To the best of our knowledge, state-of-the-art performance on the EURLEX57K dataset is 0.284 macro-F1 by LAMT\_MLC \citep{Song2022} and AttentionXML \citep{attentionxml}, and 0.762 micro-F1 by LAMT\_MLC \citep{Song2022}, as reported by \citet{Song2022}.

\subsection{Sampling Strategy}
\label{appendix:sampling} 
To investigate the effect of dataset size on model performance, we downsample the number of training data points $m$ to $m \in [1000, 2000, 5000, 10000]$ (this is done after label sampling). The sampling first picks one data point for each label combination (out of the $k$ labels) occurring in the original train split to ensure that each label combination is also represented in the sampled training set\footnote{If the number of label combinations is $> m$, not all label combinations will be included in the sampled training set.}. The remaining data points are sampled randomly. We ensure that smaller sampled training sets are sub-sets of larger ones, i.e. the training set with $1000$ data points is contained in the training set with $2000$ data points. This ensures that model performance differences observed on sampled training sets with different sizes are indeed due to the size rather than the exact data points included in the sets.

Note that for comparability we use the original dev and test sets when experimenting with the sampled training sets. However, since we do not aim to evaluate zero-shot learning here, we limit the dev and test sets to the $k$ most frequent labels.

\subsection{Sparse vector similarity-ranking}
\label{appendix:sparse}
ClassTFIDF is a similarity-based approach, where a TF-IDF vector is obtained for each label (or "class") from the concatenation of all documents in the training set with that label. To obtain predictions for a new document, the cosine-similarity between the new document's TF-IDF vector and each of the label vectors is calculated. Labels whose similarity is over a pre-defined threshold are assigned to the new document. After initial experimentation we assign the top N most similar labels instead of using a similarity threshold, where N is the median number of labels for each data point in the training set.

DocTFIDF refers to the approach where TF-IDF vectors for each document in the training set are calculated and then compared the TF-IDF vector of a new test document using cosine similarity. The new document gets assigned the set of labels that belong to the nearest training data point.

Finally, BM25 is the classic ranking function that ranks a set of documents based on the query terms appearing in each document, where the query terms is the BM25 index of the test document \citep{bm25}. Labels are then assigned as described for the DocTFIDF approach. Again, we assign the top N labels, where N is the median number of labels per document in the training set.

\subsection{Hyperparameters}
\label{appendix:hps}
We use the following hyperparameters for model training:
\begin{itemize}
    \item DistilRoBERTa
    \begin{itemize}
        \item Huggingface model: \url{https://huggingface.co/distilroberta-base}
        \item Epochs: 35
        \item Batch size: 64
        \item Early stopping patience: 8
        \item Learning rate: 0.0001
        \item Learning rate scheduler: cosine
        \item Warm-up Ratio: 0.1
    \end{itemize}
    \item LegalBERT
    \begin{itemize}
        \item Huggingface model: \url{https://huggingface.co/casehold/custom-legalbert}
        \item Epochs: 35
        \item Batch size: 64
        \item Early stopping patience: 8
        \item Learning rate: 0.0001
        \item Learning rate scheduler: cosine
        \item Warm-up Ratio: 0.1
    \end{itemize}
    \item T5
    \begin{itemize}
        \item Huggingface model: \url{https://huggingface.co/t5-small}
        \item Task prefix: "summarize:"
        \item Epochs: 20
        \item Batch size: 16
        \item Early stopping patience: 3
        \item Learning rate: 1e-4
        \item Weight Decay: 0.01
    \end{itemize}
    \item CrossEncoder
    \begin{itemize}
        \item Huggingface model: \url{https://huggingface.co/casehold/custom-legalbert}
        \item Epochs: 1
        \item Batch size: 8
        \item Number of negative samples created for each positive sample: 10
        \item The top N labels with highest predictions, where N is the median number of labels per document in the training set, are assigned at inference time
    \end{itemize}
    \item BiEncoder
    \begin{itemize}
        \item Huggingface model: \url{https://huggingface.co/casehold/custom-legalbert}
        \item Epochs: 1
        \item Batch size: 8
        \item Loss: CosineSimilarityLoss
        \item The top N labels with highest predictions, where N is the median number of labels per document in the training set, are assigned at inference time
    \end{itemize}
\end{itemize}

\subsection{Supplementary Figures}
\label{appendix:EURLEX57K}
\begin{figure*}[ht]
    \centering
    \includegraphics[width=\textwidth]{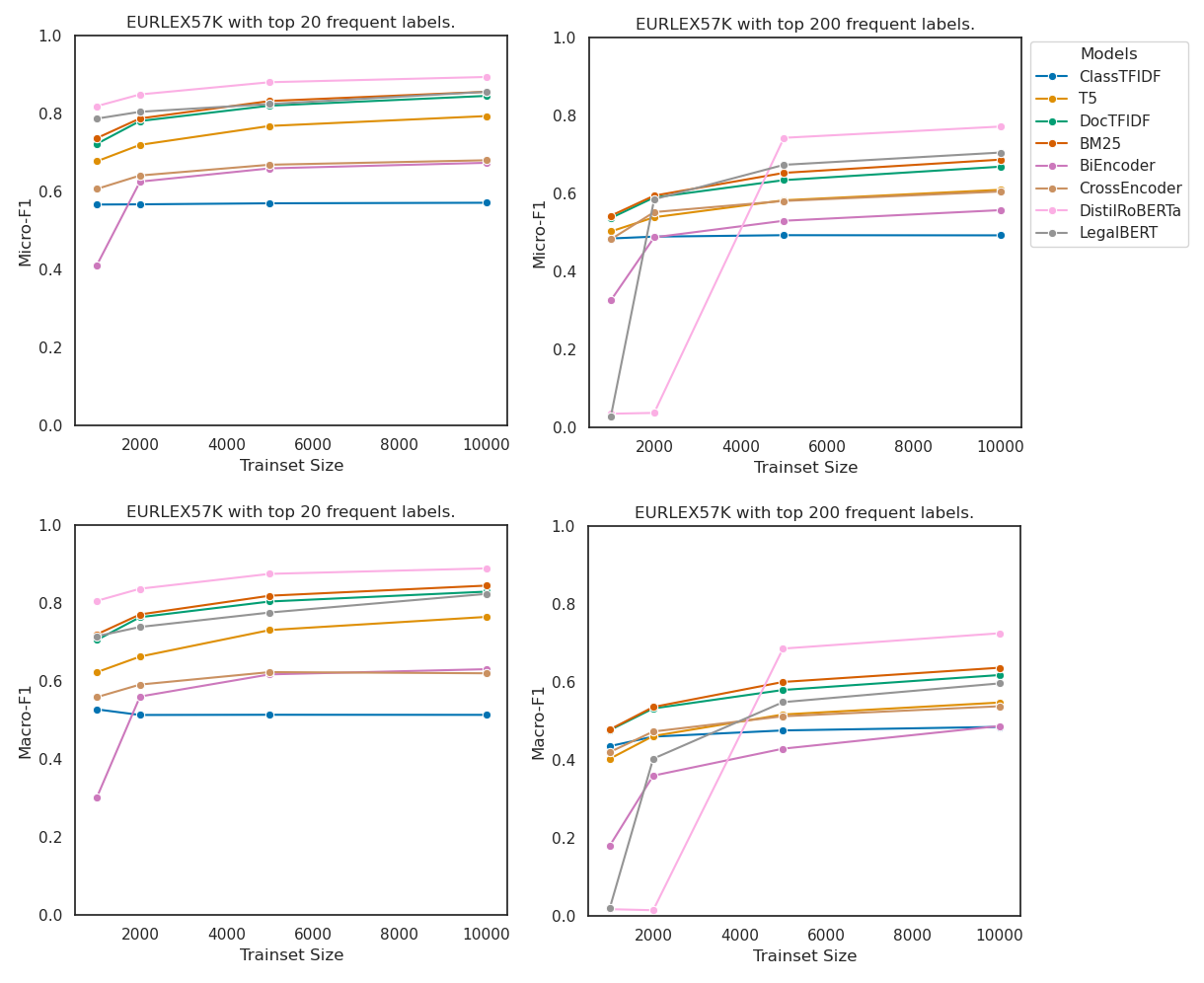}
    \caption{Micro- and macro-F1 scores of multi-label classifiers on EURLEX57K data with top 20 and top 200 labels for different training set sizes.}
    \label{fig:eurlex57k}
\end{figure*}

\begin{figure*}[ht]
    \centering
    \includegraphics[width=\textwidth]{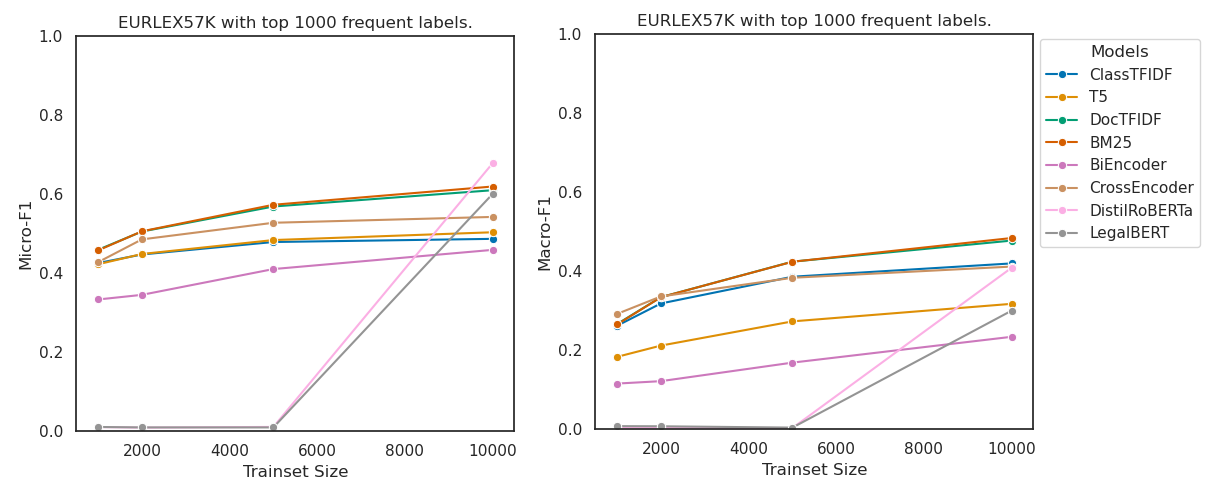}
    \caption{Micro- and macro-F1 scores of multi-label classifiers on EURLEX57K data with top 1000 labels for different training set sizes.}
    \label{fig:eurlex1000}
\end{figure*}

\begin{figure*}[ht]
    \centering
    \includegraphics[width=\textwidth]{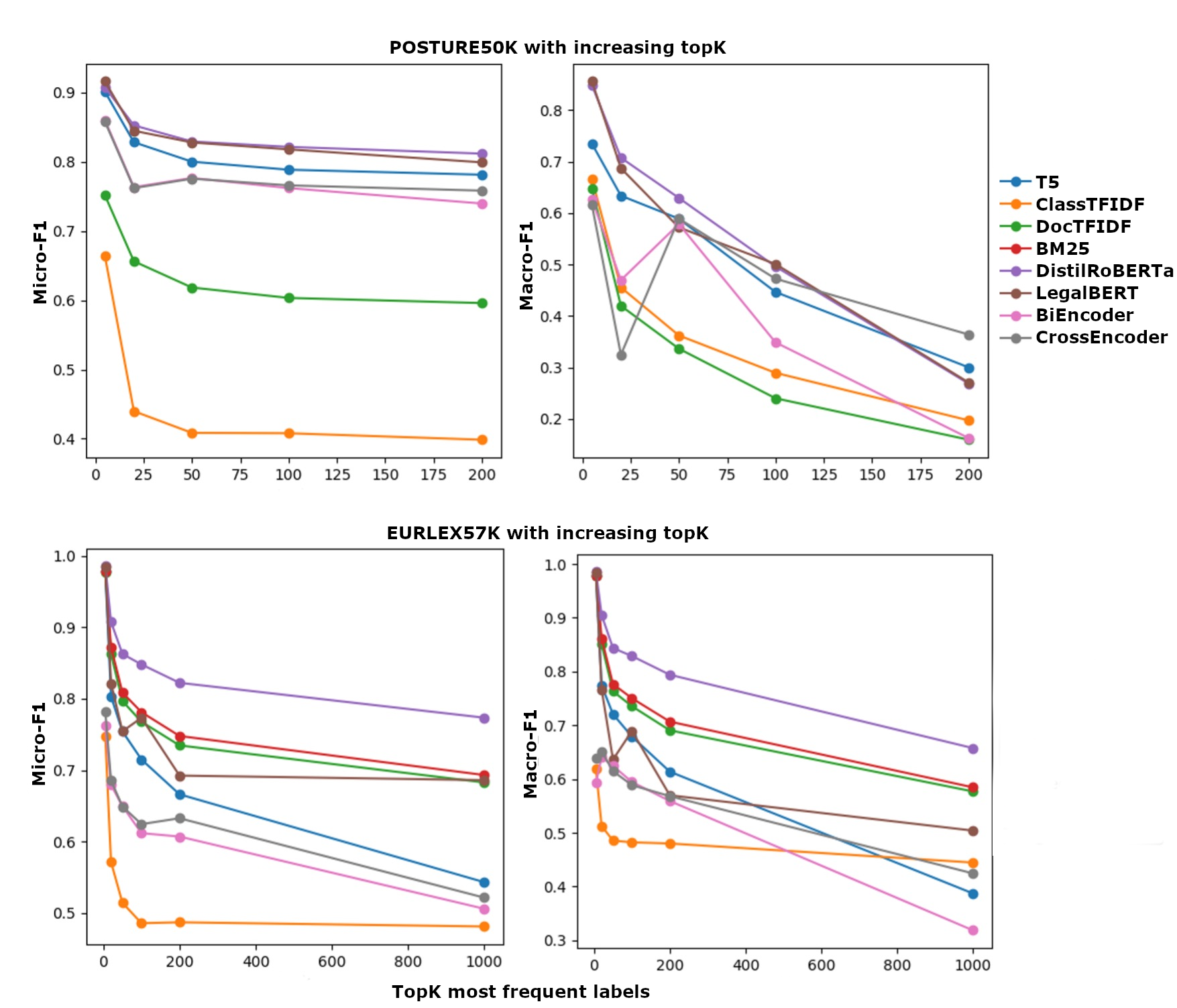}
    \caption{Performance of multi-label classifiers on the full POSTURE50K (31,944 data points) and EURLEX57K (45,000 data points) data with varying label quantities.}
    \label{fig:vary_labels}
\end{figure*}


\end{document}